\documentclass{article}

\usepackage{microtype}
\usepackage{graphicx}
\usepackage{subfigure}
\usepackage{booktabs} 
\usepackage{microtype}
\usepackage{graphicx}
\usepackage{booktabs} 
\usepackage{float}
\usepackage{placeins}
\usepackage{caption}
\usepackage{tabularx}
\usepackage{multirow}
\usepackage{multicol}
\usepackage{makecell}
\usepackage{algorithm}
\usepackage{algpseudocode} 
\usepackage{amsmath}
\usepackage{amssymb}
\usepackage{pifont}
\usepackage{mathtools}
\usepackage{amsthm}
\usepackage{xspace}
\usepackage{xcolor}

\theoremstyle{plain}

\theoremstyle{definition}

\theoremstyle{remark}

\definecolor{darkred}{RGB}{150,0,0}
\definecolor{darkgreen}{RGB}{0,150,0}
\definecolor{darkblue}{RGB}{0,0,150}
\usepackage{graphicx,bm,epsfig,epsf,color,mathbbol,fmtcount,semtrans,multirow,comment,boldline,pgfplots}
\usepackage{tikz}
\usetikzlibrary{pgfplots.groupplots}
\usepackage[utf8]{inputenc} 
\usepackage[T1]{fontenc}    
\usepackage{url}            
\usepackage{amsfonts}       
\usepackage{nicefrac}       

\newcommand{\PLUTO}{PLUTO\xspace}

\DeclareMathOperator*{\argmax}{arg\,max} 

\newcommand{\cmark}{\ding{51}}%
\newcommand{\xmark}{\ding{55}}%

\newcommand{\beq}{\begin{equation}}
\newcommand{\ba}{\begin{align}}
\newcommand{\ea}{\end{align}}

\newcommand{\eeq}{\end{equation}}

\newcommand{\x}{\vct{x}}

\definecolor{emmanuel}{RGB}{255,127,0}

\newcommand{\vct}[1]{\bm{#1}}

\usepackage{hyperref}
\usepackage{xcolor}

\usepackage[accepted]{icml2024}

\usepackage{amsmath}
\usepackage{amssymb}
\usepackage{mathtools}
\usepackage{amsthm}

\usepackage[capitalize,noabbrev]{cleveref}

\usepackage[textsize=tiny]{todonotes}

\definecolor{mymauve}{rgb}{0.58,0,0.82}

\icmltitlerunning{Plug-and-Play Transformer Modules for Test-Time Adaptation}

\begin{document}

\twocolumn[
\icmltitle{Plug-and-Play Transformer Modules for Test-Time Adaptation}



\icmlsetsymbol{equal}{*}

\begin{icmlauthorlist}
\icmlauthor{Xiangyu Chang}{ucr}
\icmlauthor{Sk Miraj Ahmed}{ucr}
\icmlauthor{Basak Guler}{ucr}
\icmlauthor{Srikanth V. Krishnamurthy}{ucr}
\icmlauthor{Ananthram Swami}{dod}
\icmlauthor{Samet Oymak}{umich}
\icmlauthor{Amit K. Roy-Chowdhury}{ucr}
\end{icmlauthorlist}

\icmlaffiliation{ucr}{University of California, Riverside}
\icmlaffiliation{dod}{DEVCOM Army Research Laboratory}
\icmlaffiliation{umich}{University of Michigan}
\icmlkeywords{Machine Learning, ICML}

\vskip 0.3in
]
\footnotetext[1]{University of California, Riverside. Email: \{cxian008@, sahme047@, bguler@ece., krish@cs., amitrc@ece.\}ucr.edu}
\footnotetext[2]{DEVCOM Army Research Laboratory. Email: ananthram.swami.civ@army.mil}
\footnotetext[3]{University of Michigan. Email: oymak@umich.edu}




\begin{abstract}


Parameter-efficient tuning (PET) methods such as LoRA, Adapter, and Visual Prompt Tuning (VPT) have found success in enabling adaptation to new domains by tuning small \emph{modules} within a transformer model. 
However, the number of domains encountered during test time can be very large, and the data is usually unlabeled. Thus, adaptation to new domains is challenging; it is also impractical to generate customized tuned modules for each such domain. 
Toward addressing these challenges, this work introduces \PLUTO: a \underline{P}lug-and-p\underline{L}ay mod\underline{U}lar \underline{T}est-time adaptati\underline{O}n strategy. We pre-train a large set of modules, each specialized for different source domains, effectively creating a ``module store''. Given a novel target domain with few-shot unlabeled data, we introduce an unsupervised test-time adaptation (TTA) method to (1) select a sparse subset of relevant modules from this store and (2) create a weighted combination of selected modules without tuning their weights. This plug-and-play nature enables us to harness multiple most-relevant source domains in a single inference call. Comprehensive evaluations demonstrate that \emph{\PLUTO uniformly outperforms alternative TTA methods} across several benchmarks and that selecting $\leq$5 modules suffices to extract most of its benefits. At a high level, our method equips pre-trained transformers with the capability to dynamically adapt to new domains, spearheading a new framework for efficient and scalable domain adaptation.


\end{abstract}
\section{Introduction}
\begin{table}[t]
  \centering
      \caption{Our work compared to the prior art fulfills key criteria for a comprehensive adaptation framework. We adapt with a sample-specific setup and can support few/zero-shot application.}
  \resizebox{\columnwidth}{!}{%
    \begin{tabular}{c cccc}
      \toprule
      Setting & \makecell{Source\\Free}  & \makecell{Adaptation\\On the fly}  & \makecell{Dynamic\\Target}  & \makecell{Few/Zero-\\Shot} \\
      \cmidrule(lr){1-5} 
      UDA & \xmark & \xmark  & \xmark   & \xmark\\
      Source-free UDA  & \cmark & \xmark  &  \xmark  & \xmark\\
      TTA  & \cmark& \cmark& \cmark& \xmark\\
      \PLUTO  &\cmark & \cmark & \cmark & \cmark\\
      \bottomrule
    \end{tabular}

  }
  \centering
\end{table}

Pretrained transformers \cite{devlin2018bert,brown2020language,liu2019roberta,radford2019language,raffel2020exploring} have achieved significant recent success, however, these models are often quite large, with billions of parameters \cite{brown2020language,raffel2020exploring}. This has posed challenges for their practical use, especially in edge devices, due to the computational and memory challenges of fine-tuning for downstream tasks. To address this, parameter-efficient tuning (PET) methods have been proposed including prefix/prompt-tuning \cite{li2021prefix,lester2021power,liu2023gpt}, adapter \cite{houlsby2019parameter}, and LoRA \cite{hu2021lora}. In PET, the pretrained model remains frozen and only the domain-specific (extra) parameters designated for the target domain are updated based on the training data. PET has found notable success as it significantly reduces memory usage and facilitates modular adaptation of the model while often being competitive with fine-tuning.

\begin{figure*}[t]
  \centering
  \includegraphics[width=0.85\textwidth]{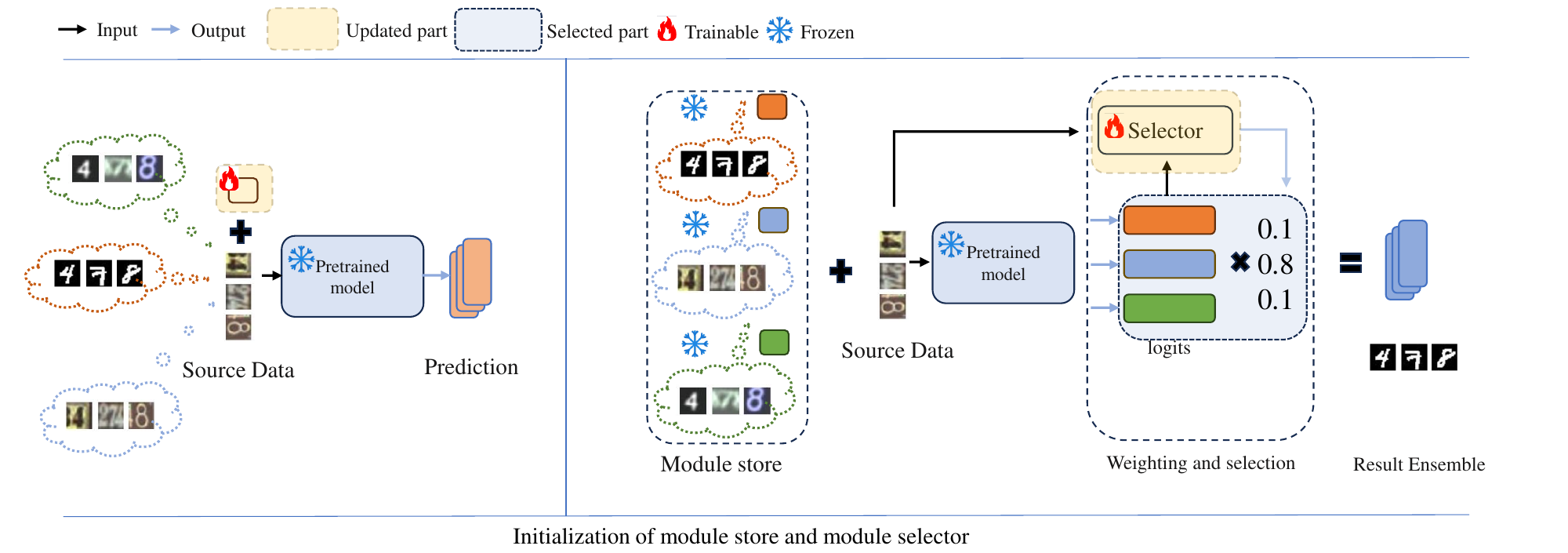}
  \includegraphics[width=0.85\textwidth]{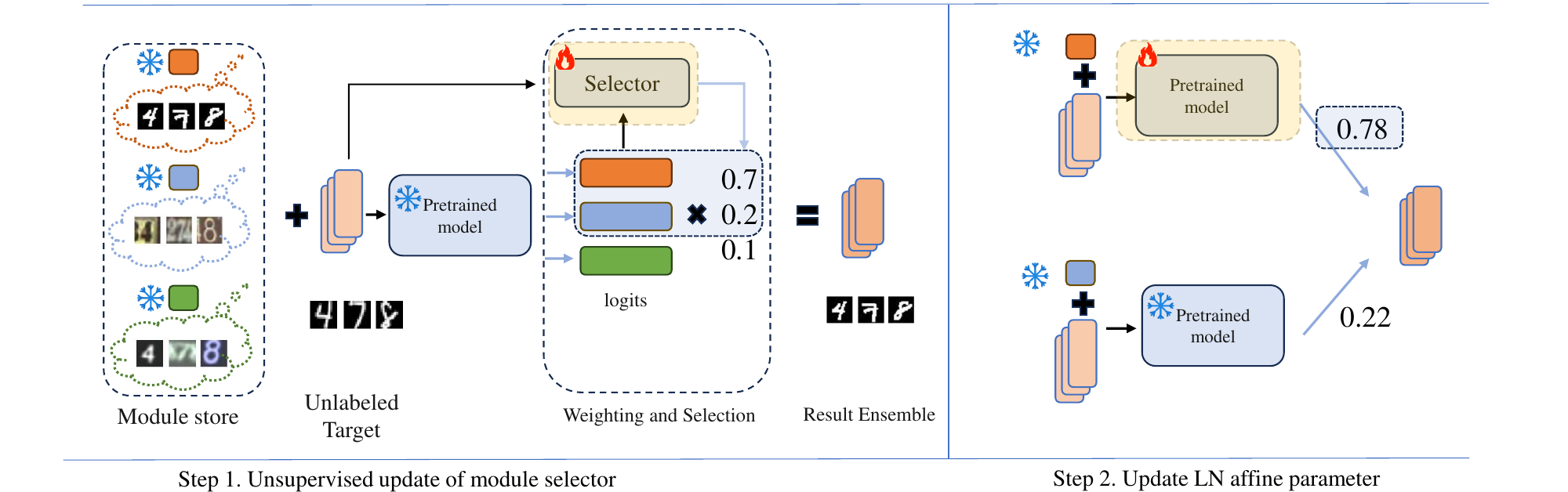}
  \caption{The overview of PLUTO. At test time, PLUTO efficiently combines the sources using appropriate weights determined by the current test distribution. Furthermore, we selectively update the LayerNorm (LN) parameters of the model that demonstrates the highest correlation with the test distribution. The numbers in the figure are examples provided for illustrative purposes.}
\label{fig:overview}
\end{figure*}
While PET methods have demonstrated notable benefits across various domains and tasks \cite{liu2023gpt,vu2021spot}, their effectiveness in more challenging scenarios involving few-shot and unlabeled data or domain shifts remains a major bottleneck \cite{gu2021ppt, raffel2020exploring}. Specifically, we ask: \emph{How to leverage PET methods to adapt to a new domain at test-time \cite{wang2020tent} where we only observe \textbf{few-shot unlabeled data} from this new domain?} Here, test-time adaptation (TTA) refers to adapting to new domains during testing, in an online manner, without the need for additional training data. This makes this setting more challenging than standard unsupervised domain adaptation (UDA) \cite{tsai2018learning,tzeng2017adversarial}. TTA typically requires iteratively updating the source model parameters using an unsupervised objective that incorporates the new test sample from the target distribution. Existing test-time adaptation (TTA) techniques mostly depend on a single source model for adaptation \cite{wang2020tent,wang2022continual}. However, considering the dynamic nature of the test distribution, it should prove advantageous to adapt a set of source models collectively during test time \cite{ye2023multi, sun2015survey}.

In this work, rather than employing multiple distinct source models, we propose to plug source module(s) within a single vision transformer. 
We naturally cannot expect a single module to work for all target domains. Furthermore, since target domains at test time are few-shot and there can be a large number of domains, it is impractical to train customized PET modules for each target from scratch. Instead, the ideal algorithm should identify the new domain and suitably adapt pre-existing PET modules to this domain. The main contribution of this work is providing a plug-and-play algorithm to address this challenge. We call our method \textbf{\PLUTO: \underline{P}lug-and-p\underline{L}ay mod\underline{U}lar \underline{T}est-time adaptati\underline{O}n}.


Concretely, we consider a multi-source domain adaptation setting and propose \emph{pretraining PET modules for a collection of source domains}. These pretrained modules can then be utilized at test-time with minimal supervision and sample size. Importantly, our algorithm \PLUTO benefits from having access to a \emph{module store} containing a diverse set of source domains. \PLUTO can retrieve the most relevant sources from this \emph{module store} (70+ modules for ImageNet-C benchmark) and accurately blend them in a target-aware fashion.
    \PLUTO brings three important benefits: 
    {(1) {\em Sample efficiency}:} \PLUTO enables few-shot unsupervised adaptation in two ways: First, \PLUTO only selects and weights a small subset of pretrained modules, rather than tuning a module from scratch, resulting in an efficient search space. Secondly, we utilize an attention-based module selector to identify the most relevant modules to the target domain. This also allows \PLUTO to exhibit strong zero-shot adaptation performance.
    {(2) {\em Anti-forgetting during TTA}: } Our plug-and-play use of PET modules and selective update of LayerNorm parameters effectively overcomes the forgetting which is a prominent challenge for TTA methods (see Section.~\ref{sec:antiforget} for details).
    {(3) {\em Parameter efficiency}:} The adaptation happens through parameter-efficient modules. In a distributed/federated setting where central server operates a \emph{module store}, the clients can efficiently request new modules.

\noindent\textbf{Main Contributions.} Overall, our method \PLUTO is an effective strategy for harnessing not only a pretrained Vision Transformer (ViT) but also pretrained PET modules within this ViT to enhance few-shot unsupervised TTA performance. Our main technical contributions are:\vspace{-10pt}
\begin{itemize}
    \item \textbf{Multi-source PET-based TTA.} To the best of our knowledge, this is the first work to employ multiple PET modules for the test-time adaptation problem (hence the phrase \emph{plug-and-play}). We not only pretrain a large number of modules, we also present innovative algorithms that can effectively select and blend an optimal subset of these modules to facilitate few-shot unsupervised adaptation to a new domain. Importantly, \PLUTO achieves a performance on par with using all modules simultaneously and often outperforms the single best-performing source in hindsight.

    \item \textbf{Algorithmic innovations.} \PLUTO facilitates multi-source TTA by weighting the output logits of multiple modules and fine-tuning LayerNorms (LN). Our method crucially captures the relationships between the logit outputs of multiple source modules and input instances with very limited samples and enables \PLUTO to decide the contribution of each source domain adaptively on the fly. To ensure the generalizability of the few-shot weight assignment of the source modules, we integrate sharpness-aware minimization and PET methods \cite{foret2020sharpness} for TTA.
    \item \textbf{Empirical impact and insights.} Comprehensive evaluations on Digits, Office-Home, CIFAR-10C, and ImageNet-C demonstrate that \PLUTO uniformly outperforms alternative TTA methods. The performance improvement is much larger in the few-shot setting. For instance, in Office-Home evaluations in Table \ref{table:fewshot_OfficeHome}, \PLUTO demonstrate \textbf{+24\% absolute accuracy improvement in zero-shot and 16-shot settings} over the best alternative. To do so, \PLUTO typically selects only $\leq$5 modules to extract most of the benefits.
\end{itemize}
\section{Related Works}
\subsection{Unsupervised Domain Adaptation}

Unsupervised Domain Adaptation (UDA) is widely used in various machine learning tasks, including image classification \cite{tzeng2017adversarial}, semantic segmentation \cite{tsai2018learning}, object detection \cite{hsu2020progressive}, and reinforcement learning \cite{raychaudhuri2021cross}. These applications address distribution shift challenges by harmonizing source and target domain distributions. Methods utilize techniques like maximum mean discrepancy and adversarial learning \cite{ganin2016domain,tzeng2017adversarial} for alignment. There is a growing interest in adaptation strategies that rely solely on pretrained source models due to privacy and memory storage concerns regarding source data.

\subsection{Test Time Adaptation (TTA)} 
UDA requires extensive target domain data for offline adaptation, whereas TTA employs a continual, batch-by-batch approach \cite{valanarasu2022fly,shin2022mm,hu2021fully}. Initial studies \cite{li2016revisiting} adapted using the current test-batch statistics rather than those from training. TENT \cite{wang2020tent} aims to reduce entropy and updates batch-normalization parameters in a pre-trained source model using new target data. DUA \cite{mirza2022norm} persistently updates batch-norm statistics with incoming test batches for better target domain alignment. A common issue with these approaches is the tendency to forget source knowledge. Methods like CoTTA \cite{wang2022continual} apply stochastic source model restoration to avoid drifting of source knowledge, and EATA \cite{niu2022efficient} uses regularization to preserve important weights, thus reducing forgetting. In addition, CLIP~\cite{radford2021learning}-based methods \cite{feng2023diverse,shu2022test} adapt to new domains by doing data augmentation and prompt generation. These single-source TTA methods require substantial computational resources to balance the trade-off between adaptation performance and mitigating forgetting. In contrast, by assessing the relevance between multiple source models and the target domain and only updating the most related source models, our approach minimizes forgetting with minimal computational expense while maximizing the utilization of knowledge from all source domains.

\subsection{Ensemble learning} Ensemble learning, a well-established approach, aims to create a robust and generalizable model \cite{hansen1990neural}. Common ensemble techniques include voting~\cite{hansen1990neural}, bagging~\cite{breiman1996bagging}, boosting~\cite{schapire1990strength,freund1997decision}, and stacking~\cite{wolpert1992stacked}. These methods have found applications in various tasks, including model debiasing~\cite{elazar2018adversarial,stacey2020avoiding},  and domain adaptation~\cite{kim2017domain}.
Apart from considering global domain-to-domain relationships, \cite{ruder2017learning, guo2018multi} introduced the use of similarity measures between examples and domains to choose data from various sources for a particular target domain. These selected data are then merged to train a single classifier. Some closely related works to ours are~\cite{kim2017domain,SESoM}, which also incorporate example-to-domain relationships but utilize an attention mechanism. They train the attention module using limited labeled data from the target domain in a supervised manner. In contrast, our method operates in an unsupervised setting and does not rely on such labeled data.

\vspace{-5pt}\section{Problem Formulation and Proposed Method}
We assume access to a pretrained transformer model $f$. We will build a set of pretrained modules, referred to as \emph{module store}, that can be plugged within $f$ on demand. Let us denote these modules by parameters $\{\theta_j\}_{j=1}^N$ where each $\theta_j$ will be trained for a particular source domain. These modules, based on parameter-efficient tuning methods, will contain much fewer parameters than the base model $f$. Let us also denote the transformer $f$ with module $\theta$ as $f_\theta$. Before describing \PLUTO, we introduce our workflow: We will (1) pretrain source modules $\{\theta_j\}_{j=1}^N$ using source datasets and (2) optimally blend these modules for a target task.\vspace{-5pt}

\begin{itemize}
    \item \textbf{Pretraining Source Modules:} Given a loss function $\mathcal{L}$ and a fixed set of source domains $\{S_j\}_{j=1}^N$, for each $\mathcal{S}\in\{S_j\}_{j=1}^N$, instead of fully fine-tuning ($f\rightarrow f^{\mathcal{S}}$) the pretrained model to the domain $\mathcal{S}$ by solving $f^{\mathcal{S}}=\text{argmin}_f \mathcal{L}(f;\mathcal{S})$, given the pretrained model (with modules) $f_\theta$, we freeze the model and only update the module $\theta=\text{argmin}_\theta \mathcal{L}(f_{\theta};\mathcal{S})$. \vspace{-5pt}

    \item \textbf{Adapt to a new domain:} After obtaining a set of $N$ modules $\{\theta_j\}_{j=1}^N$ corresponding to $\{\mathcal{S}_j\}_{j=1}^N$, our objective is to adapt to a new target domain $\mathcal{T}_{new}$ without fine-tuning a new module but only utilizing the current pretrained modules. We aim to find a combination $\mathcal{G}$ of the source models $\mathcal{G}(\{f_{\theta_j}\}_{j=1}^N)$ that minimizes the loss on the new domain:
    \[
    \text{min}_\mathcal{G} \mathcal{L}(\mathcal{G}(\{f_{\theta_j}\}_{j=1}^N);\mathcal{T}_{new}).
    \]
    Specifically, $\mathcal{G}$ can be a linear combination of the source models' output. This would apply under the assumption that the target domain distribution can be expressed as the linear combination of source distributions. We'll provide some theoretical insights of this approximation in Appendix \ref{sec:theory}. 
\end{itemize}\vspace{-5pt}

\noindent\textbf{Method:} To proceed, we provide an overview of \PLUTO's framework illustrated in Figure~\ref{fig:overview}. This involves the following steps to achieve unlabeled test-time domain adaptation:
\begin{itemize}

    \item \textbf{(Sec.~\ref{sec:module_selection}) Step 1:} For a test batch at time $t$, we have both the pretrained modules and the frozen pretrained transformer model. We update the \emph{module selector} $\mathcal{G}$ with pseudo-label entropy minimization. This process helps us learn the module output combination weights and select the most related source modules specific to the current batch. As we discuss in Sections \ref{sec:initialization} and \ref{sec:module_selection}, \emph{module selector} $\mathcal{G}$ relies on an attention mechanism and plays a crucial role in effectively determining the source modules that are most relevant to the target.
    
    \item \textbf{(Sec.~\ref{sec:LN_adaptation}) Step 2:} At each time $t$, we determine which source model is needed to update the LayerNorm (LN) affine parameters by the largest assigned weight from the updated module selector $\mathcal{G}$. 
\end{itemize}
    In both steps, we only require a small number of unlabeled test samples to identify the source modules that are most relevant to the current test batch. We can typically select just a few modules ($\leq$5) to achieve performance that approximates the effect of using all weighted source module outputs during forward inference. The pseudocode for \PLUTO is provided in Algorithm~\ref{alg:module_store}. In what follows, we further elaborate on the design principles of \PLUTO.

\subsection{Initialization of the Module Selector}
\label{sec:initialization}
Given the correspondence between $N$ source domains $S_1,\dots,S_N$ and their  labeled training data, by running parameter-efficient tuning on each domain separately, a set of pretrained modules, or the \emph{module store}, can be obtained: $\{\theta_j\}_{j=1}^N$. Given a certain pretrained transformer model $f$, we denote the model equipped with different pretrained source modules as $\{f_{\theta_j}^j\}_{j=1}^N$.  
  
For source domains $\{S_j\}_{j=1}^N$ the ground-truth (GT) label of the target sample $x$ is known, and the module selector $\mathcal{G}$ can be initialized by minimizing the cross-entropy loss between the weighted label prediction (will be explained in Sec.~\ref{sec:module_selection}) and the GT label $y$. As discussed in \cite{SESoM}, this initialization enables $\mathcal{G}$ to capture the sample-specific preference of different source models.

\begin{algorithm}[t]
\caption{PLUTO: \underline{P}lug-and-p\underline{L}ay mod\underline{U}lar \underline{T}est-time adaptati\underline{O}n}
\label{alg:module_store}
\begin{algorithmic}[1]
\State \textbf{Input:} Source model with pretrained source modules $\{f_{\theta_j}^j\}_{j=1}^N$, attention-based pretrained module selector $\mathcal{G}$,  number of modules to be selected $M$ ($M<N$), streaming sequential unlabeled test data $\{x_i^{(1)}\}_{i=1}^B\rightarrow \{x_i^{(2)}\}_{i=1}^B\rightarrow \dots \{x_i^{(t)}\}_{i=1}^B\rightarrow \dots$
\State \textbf{Output: } $M$ scaled weights, $M$ selected modules, Updated LayerNorm (LN) affine parameters
\While{$t \ge 1$}
        \For{Each $x_i$ in the $t$-th batch}
        \State Pass $x_i$ through $\{f_{\theta_j}^j\}_{j=1}^N$ 
        \State Obtain pre-softmax logits $\{l(x_i)^j\}_{j=1}^N$
        \State Obtain weights 
        \Statex \hspace{\algorithmicindent} $\{w(x_i)^j\}_{j=1}^N=\mathcal{G}(x_i,\{l(x_i)^j\}_{j=1}^N)$
        \State Assign pseudo-label (PL) to $x_i$
        \EndFor
    \State Update $\mathcal{G}\rightarrow \mathcal{G^*}$ with PL entropy minimization {on $t$-th batch} (Eqn.~\ref{eq:PLentropy})
    \State Obtain updated weights for all $x_i$ in $t$-th batch
    \Statex \hspace{\algorithmicindent} $\forall x_i, \{w^*(x_i)^j\}_{j=1}^N=\mathcal{G^*}(x_i,\{l(x_i)^j\}_{j=1}^N)$
    \State Calculate average weights for $t$-th batch 
    \Statex \hspace{\algorithmicindent} $\{\overline{w}(t)^j\}_{j=1}^N=\{\overline{\{w^*(x_i)^j\}_{i=1}^B}\}_{j=1}^N$
    \State Find and select the top $M$ in $\{\overline{w}(t)^j\}_{j=1}^N$
    \State Rescale selected weights to sum up to 1.
    \State LN update of source model corresponding to the largest in $\{\overline{w}(t)^j\}_{j=1}^N$ by minimizing Eqn.~\ref{eq:LNentropy} with Eqn.~\ref{eq:SAM_GradientApprox}
\EndWhile
\end{algorithmic}
\end{algorithm}

\subsection{Module weighting at test time}
\label{sec:module_selection}
Drawing from the success of the attention mechanism in achieving impressive few-shot performance in supervised model ensemble scenarios \cite{SESoM, kim2017domain}, we design a novel approach to leverage the same during test time adaptation. Even in the unsupervised context of test-time adaptation, we have found that an attention module, which acts as a module selector, can effectively integrate information from multiple source domains into the target domain by dynamically assigning weights and updating them on-the-fly. This enables \PLUTO to plug modules from the module store and adapt them at test time with few-shot unlabeled data.

In order to convert a 2D image input into sequential data for the transformer and the module selector to process, the image input $x\in \mathbb{R}^{H\times W\times C}$ is reshaped into a sequence of flattened 2D patches \cite{dosovitskiy2020image}~$\mathbf{X}_p\in \mathbb{R}^{e\times (P^2\cdot C)}$, where $(H,W)$ is the height and width of the original image, $C$ is the number of channels, $(P,P)$ is the size of each segmented image patch, and $e = HW/P^2$ is number of patches and the effective input sequence l\underline{\textbf{e}}ngth. Assuming that the transformer uses a constant latent vector size $d$ across all of its layers\cite{dosovitskiy2020image}, we flatten the sequence of the patches and map them to $d$ dimentions: $\hat{\mathbf{X}}=[ \mathbf{X}_p\mathbf{E} ]_{p=1}^N, \mathbf{E}\in\mathbb{R}^{(P^2\cdot C)\times d}$. The patch embedding projection $\mathbf{E}$ is a (frozen) trainable parameter in the pretrained ViT.

The module selector, which is an attention module, gauges the relevance between the output logits of the source modules and instances $\x_i$ in the current ($t$-th) \textbf{t}arget (abbreviated as $T$) batch $D_T^{(t)}=\{x_i^{(t)}\}_{i=1}^B$. 
It then assigns weights based on this relevance. Logits of source models are a function of both the pretrained module and the pretrained transformer and are also capable of modeling the uncertainty in predictions from different sources.
We perform max-pooling over the sequence $\hat{\mathbf{X}}=[ \mathbf{X}_p\mathbf{E} ]_{p=1}^N \in \mathbb{R}^{e\times d}$ and obtain $\hat{x}\in \mathbb{R}^d$ as the representation of the original image input. Furthermore, we pass the input instance $x$ through $N$ pretrained models and obtain pre-softmax logits $\{l(x_i)^j\}_{j=1}^N\in \mathbb{R}^v$. With four trainable parameters:
$W_{d,x}\in \mathbb{R}^{d\times d'_x}, W_{u,x}\in \mathbb{R}^{d'_x\times d'}, W_{d,l}\in \mathbb{R}^{v\times d'_l}, W_{u,l}\in \mathbb{R}^{d'_l\times d'}$, the module selector first projects the pre-softmax logits and the representation of the original image input into another representational space: $\mathbf{h}_x = W_{u,x}^\intercal\cdot \gamma(W_{d,x}^\intercal\cdot \hat{x}), \mathbf{h}_{l,j} = W_{u,l}^\intercal\cdot \gamma(W_{d,l}^\intercal\cdot {l(x)^j})$. With application of LN \cite{ba2016layer} we are able to obtain the final projected representations $\mathbf{h}_x ,\mathbf{h}_{l,j}\in \mathbf{R}^{d'}$ and compute the attention weight $w(x)^j$ to capture the relationship between the input $x$ and its representation from the pretrained models $l(x)^j$:
\begin{align}
w(x)^j=\mathcal{G}(x_i,\{l(x_i)^j\}_{j=1}^N)=\frac{e^{\mathbf{h}_{l,j} \cdot \mathbf{h}_x}}{\sum_{k=1}^N e^{\mathbf{h}_{l,k} \cdot \mathbf{h}_x}}
\label{eq:attn_score}
\end{align}

The ensembled logit is the linear combination of the weighted logits:
$l(x)=\sum_{j=1}^N w(x)^j l(x)^j.$

As the GT label of the target sample $x$ is unknown, we denote its pseudo-label, as predicted by source $j$, as $\hat{y}_j=f^j_{\theta_j}(x)$. We linearly combine these pseudo-labels by attention weights $\{w(x)^j\}_{j=1}^N$ to get the weighted pseudo-label $\hat{y}=\sum_{j=1}^N w(x)^j \hat{y}_j$. Using these weighted pseudo-labels for all the samples in the current ($t$-th) batch, we calculate the expected Shannon entropy of pseudo-label as: 
\begin{align}
\mathcal{L}=-\mathbf{E}_{\mathcal{D}_T^{(t)}}\sum_{c=1}^K \hat{y}_c\log(\hat{y}_c),
\label{eq:PLentropy}
\end{align}
where $K$ is the number of classes and $\hat{y}_c$ is the $c$-th entry of the predicted pseudo-label. By performing entropy minimization on Eqn.~\ref{eq:PLentropy}, we can update the module selector parameters (from $\mathcal{G}$ to $\mathcal{G}^*$) and obtain updated attention weights for Eqn.~\ref{eq:attn_score}. Minimization of the Shannon entropy attempts to raise the confidence of an individual sample's prediction and weight assignment \cite{lee2023towards}. 

\subsection{Domain adaptation of LN affine parameters}
\label{sec:LN_adaptation}
In TTA, prior methods often conduct adaptation on
pre-trained models with batch normalization (BN) layers~\cite{ioffe2015batch}  and most of them are built upon BN statistics adaptation~\cite{schneider2020improving,wang2020tent,zhang2022memo}. However, recent works~\cite{liu2022convnet,de2023effectiveness,kim2021adapt} found out that layer normalization (LN, \cite{ba2016layer}) can be a better adaptation choice for transformers.  Specifically, for the $k$-th layer with a $d$-dimensional input $x=(x^{(1)},\cdots,x^{(d)})^\intercal\in\mathbb{R}^d$, the layer normalized outputs are 
\begin{align}
y^{(k)}=\gamma^{(k)}\hat{x}^{(k)}+\beta^{(k)},
\text{i.e., } y=\gamma\odot\hat{x}+\beta, 
\label{eq:LNaffine}
\end{align}
where $\hat{x}^{(k)}=\frac{(x^{(k)}-\mu_d)}{\sqrt{\sigma_d^2+\epsilon_\sigma}}$, 
$\mu_d=\frac{1}{d}\sum_{k=1}^d x^{(k)},\\ \sigma_d=\frac{1}{d}\sum_{k=1}^d (x^{(k)}-\mu_d)^2$. Here, $\gamma^{(k)}$ and $\beta^{(k)}$ are learnable affine parameters (and we denote them together as $\lambda$), $\epsilon_\sigma$ is a small value to ensure numerical convergence. Adaptation to new domains can be achieved by tuning $\gamma^{(k)}$ and $\beta^{(k)}$ in Eqn.~\ref{eq:LNaffine}.

\subsubsection{Sharpness-aware domain adaptation}

While TTA provides stability in LN models, it can lead to model collapse during affine parameter tuning (Eqn.~\ref{eq:LNaffine}). This collapse occurs when the model incorrectly assigns all input samples to a single class over the adaptation process~\cite{niu2023towards}. To prevent this, we utilize sharpness-aware techniques \cite{foret2020sharpness,niu2023towards} to make the model less sensitive to large gradients and in test samples~\cite{niu2022efficient}.

Upon obtaining the updated weights from the updated module selector $\mathcal{G^*}$ for each test sample in the target batch $D_T^{(t)}=\{x_i^{(t)}\}_{i=1}^B$ at time step $t$, we identify the $M$ source modules most relevant to the target batch based on the highest average weights assigned by the module selector.
We compute the entropy of the pseudo-labels of current test batch as predicted by these models with pretrained module. The entropy of the prediction of $t$-th target batch from $j$-th source model is: 
\begin{align}
    \mathcal{L}_j^{(t)}=-\mathbf{E}_{\mathcal{D}_T^{(t)}}\sum_{c=1}^K \hat{y}_{jc}^{(t)}\log(\hat{y}_{jc}^{(t)})
\label{eq:LNentropy}
\end{align}

We seek to make the model insensitive to the large gradients by encouraging the model to converge to a flat area of the entropy loss surface, since a flat minima has good generalization and robustness to large gradients~\cite{foret2020sharpness,niu2023towards}:
\begin{align}
    &\min_\lambda \mathcal{L}_j^{SA(t)}(\{x_i^{(t)}\}_{i=1}^B;\mathbf{\lambda}), \\
    &\text{where }\mathcal{L}_j^{SA(t)}\triangleq \max_{\|\mathbf{\epsilon}\|_2\leq\rho}\mathcal{L}_j^{(t)}(\{x_i^{(t)}\}_{i=1}^B;\mathbf{\lambda}+\mathbf{\epsilon})
\label{eq:SAM}
\end{align}
The abbreviation "SA" stands for sharpness-aware. In this context, the inner optimization aims to discover a perturbation $\mathbf{\epsilon}$ of LayerNorm affine parameters $\mathbf{\lambda}$ within a Euclidean ball of radius
$\rho$ that maximizes entropy. The degree of sharpness is measured by the maximum change in the Euclidean ball neighbourhood $N_\rho(\mathbf{\lambda})$. This bi-level problem incentivizes the optimization process to locate flat minima. Following SAM~\cite{foret2020sharpness}, we can approximately solve the inner optimization via a first-order Taylor expansion,
\begin{align*}
&\mathbf{\epsilon^*}(\mathbf{\lambda}) \triangleq \argmax_{\|\mathbf{\epsilon}\|_2\leq\rho}\mathcal{L}_j^{(t)}(\{x_i^{(t)}\}_{i=1}^B;\mathbf{\lambda}+\mathbf{\epsilon}) \\
&\approx \argmax_{\|\mathbf{\epsilon}\|_2\leq\rho}\mathcal{L}_j^{(t)}(\{x_i^{(t)}\}_{i=1}^B;\mathbf{\lambda})+\mathbf{\epsilon}^\intercal\nabla_\lambda\mathcal{L}_j^{(t)}(\{x_i^{(t)}\}_{i=1}^B;\mathbf{\lambda})\\
&=\argmax_{\|\mathbf{\epsilon}\|_2\leq\rho} \mathbf{\epsilon}^\intercal\nabla_\lambda\mathcal{L}_j^{(t)}(\{x_i^{(t)}\}_{i=1}^B;\mathbf{\lambda})
\end{align*}
Denote $\mathbf{v}=\nabla_\lambda\mathcal{L}_j^{(t)}(\{x_i^{(t)}\}_{i=1}^B;\mathbf{\lambda})$. Hölder's inequality implies that 
$\epsilon^\mathbf{T}\mathbf{v}\le\|\epsilon\|_p\|\mathbf{v}\|_q\le \rho\|\mathbf{v}\|_q$ ($1/p+1/q=1$).
For $p=q=2$, the linear function achieves that bound $\mathbf{\epsilon^*}(\mathbf{\lambda}) ^\intercal\mathbf{v}=\rho\|\mathbf{v}\|_2$,
and \begin{align}
    \mathbf{\epsilon^*}(\mathbf{\lambda})=\rho \cdot\text{sgn}(\mathbf{v})\cdot\frac{|\mathbf{v}|}{\|\mathbf{v}\|_2}
\label{eq:SAMsol}
\end{align}
By substituting $\mathbf{\epsilon^*}(\mathbf{\lambda})$(Eqn.~\ref{eq:SAMsol}) back into Eqn.~\ref{eq:SAM} and differentiating both sides, the final gradient approximation is:\begin{align}
    \nabla_\lambda \mathcal{L}_j^{SA(t)}
    \approx
    \nabla_\lambda \mathcal{L}_j^{(t)}(\{x_i^{(t)}\}_{i=1}^B;\mathbf{\lambda})|_{\lambda+\mathbf{\epsilon^*(\mathbf{\lambda})}}.
    \label{eq:SAM_GradientApprox}
\end{align}
By applying Eqn.~\ref{eq:SAM_GradientApprox} to update the LN affine parameters of $j$-th model with Eqn.~\ref{eq:LNentropy}, we can obtain a more reliable solution for entropy minimization. We update the LayerNorm affine parameters for all the layers of $j$-th source model by back propagating $\mathcal{L}_j^{(t)}$ once, as the approach in \cite{wang2020tent}. After updating $\gamma$s and $\beta$s of LN layers of $j$-th model, we denote the updated model as $\overline{ f_{\theta_j}^{j(t)}}$ which will be used of the inference on $(t+1)$-th target test batch. This implies $f_{\theta_j}^{j(t+1)}=\overline{ f_{\theta_j}^{j(t)}}$.

\section{Evaluations}
\begin{table}[t]
  \centering
      \caption{\textbf{Results on Digits}. We train the source modules using 4 digits datasets to perform inference on the remaining dataset ($M=N=4$). The table clearly shows that the average accuracy of \PLUTO outperforms all of the baselines consistently.}
  \resizebox{\columnwidth}{!}{%
\begin{tabular}{ccccccc|c}
\hline
\textbf{Source}         & \textbf{Method} & \textbf{MM}   & \textbf{MT}   & \textbf{UP}   & \multicolumn{1}{l}{\textbf{SV}}   & \textbf{SY}   & \textbf{Avg}  \\ \hline
\multirow{5}{*}{Single} & TENT-Best       & 56.1          & 98.4          & 84.9          & 87.0                                & 95.2          & 84.3          \\
                        & TENT-Worst      & 17.6          & 54.2          & 59.6          & 11.4                              & 15.5          & 31.7          \\
                        & CoTTA-Best      & 55.1          & 97.5          & 84.9          & 86.3                              & 93.6          & 83.5          \\
                        & CoTTA-Worst     & 17.2          & 54.2          & 59.7          & 10.8                              & 15.6          & 31.5          \\
                        & EaTA-Best       & 54.8          & 97.3          & 84.2          & 85.4                              & 93.0            & 82.9          \\
\multicolumn{1}{l}{}    & EaTA-Worst      & 16.9          & 53.4          & 58.9          & 10.4                              & 15.8          & 31.1          \\ \hline
\multirow{4}{*}{Multi}  & TENT-Ens        & 62.4          & 97.8          & 87.6          & 55.1                              & 78.4          & 76.3          \\
                        & CoTTA-Ens       & 55.7          & 97.9          & 85.7          & 87.1                              &  \textbf{95.5}          & 84.4          \\
                        & EaTA-Ens        & 55.6          & 98.0            & 85.2          & 85.9                              & 94.9          & 83.9          \\
                        & \PLUTO (Ours)    & \textbf{63.1} & \textbf{98.9} & \textbf{89.6} & \multicolumn{1}{l}{\textbf{87.4}} & 93.7 & \textbf{86.6} \\ \hline
\end{tabular}
  }
  \centering
  \label{table:Digits}
\end{table}

\begin{table}[t]

  \centering
      \caption{\textbf{Results on Office-Home}. We train the source modules using 3 datasets to perform inference on the remaining dataset ($M=N=3$). The table clearly shows that the average accuracy of \PLUTO outperforms all of the baselines consistently.}
  \resizebox{\columnwidth}{!}{%
\begin{tabular}{cccccc|c}
\hline
\textbf{Source}         & \textbf{Method} & \textbf{Ar}   & \textbf{Cl}   & \textbf{Pr}   & \textbf{Re}   & \textbf{Avg}  \\ \hline
\multirow{5}{*}{Single} & TENT-Best       & 65.5          & 51.2          & 76.8          & 71.4          & 66.2          \\
                        & TENT-Worst      & 46.6          & 45.7          & 56.3          & 64.0            & 53.2          \\
                        & CoTTA-Best      & 64.4          & 50.2          & 76.7          & 70.5          & 65.5          \\
                        & CoTTA-Worst     & 46.6          & 46.0            & 56.0            & 63.9          & 53.1          \\
                        & EaTA-Best       & 64.8          & 50.9          & 77.1          & 71.4          & 66.1          \\
\multicolumn{1}{l}{}    & EaTA-Worst      & 47.4          & 46.7          & 55.4          & 62.8          & 53.1          \\ \hline
\multirow{4}{*}{Multi}  & TENT-Ens        & 63.1          & 49.2          & 75.6          & 71.0            & 64.7          \\
                        & CoTTA-Ens       & 64.6          & 50.4          & 76.4          & 71.7          & 65.8          \\
                        & EaTA-Ens        & 65.8          & 51.2          & 76.7          & 72.6          & 66.6          \\
                        & \PLUTO (Ours)    & \textbf{66.0} & \textbf{53.6} & \textbf{79.2} & \textbf{76.3} & \textbf{68.7} \\ \hline 
\end{tabular}
  }
  \centering
  \label{table:OfficeHome}
\end{table}

\begin{table}[t]

\centering
\caption{\textbf{Results on Digits}. Different number of few-shot target samples for training the module selector and tuning the LayerNorm affine parameters.}

  \resizebox{\columnwidth}{!}{
\begin{tabular}{ccccccc|c}
\hline
Data Size                  & Method    & MM            & MT            & UP            & SV            & SY            & Avg           \\ \hline
\multirow{4}{*}{128}       & TENT-Ens  & 62.4          & 97.8          & 87.6          & 55.1          & 78.4          & 76.3          \\
                           & CoTTA-Ens & 55.7          & 97.9          & 85.7          & 87.1          & 95.5          & 84.4          \\
                           & EaTA-Ens  & 55.6          & 98.0          & 85.2          & 85.9          & 94.9          & 83.9          \\
                           & \PLUTO     & \textbf{63.1} & \textbf{98.9} & \textbf{89.6} & \textbf{86.4} & \textbf{93.7} & \textbf{86.4} \\ \hline
\multirow{4}{*}{64}        & TENT-Ens  & 53.8          & 89.5          & 79.2          & 43.3          & 67.4          & 66.6          \\
                           & CoTTA-Ens & 45.1          & 89.9          & 77.6          & 75.3          & 87.3          & 75.0          \\
                           & EaTA-Ens  & 43.7          & 87.1          & 73.5          & 76.0          & 86.6          & 73.4          \\
                           & \PLUTO     & \textbf{59.8} & \textbf{94.6} & \textbf{86.5} & \textbf{81.8} & \textbf{88.3} & \textbf{82.2} \\ \hline
\multirow{4}{*}{16}        & TENT-Ens  & 47.8          & 84.2          & 73.0            & 37.6          & 60.5          & 60.6          \\
                           & CoTTA-Ens & 38.3          & 82.4          & 71.5          & 68.8          & 82.0          & 68.6          \\
                           & EaTA-Ens  & 38            & 81.1          & 67.7          & 70.3          & 81.6          & 67.7          \\
                           & \PLUTO     & \textbf{55.6} & \textbf{90.2} & \textbf{81.0} & \textbf{77.0} & \textbf{83.1} & \textbf{77.4} \\ \hline
\multirow{4}{*}{Zero-shot} & TENT-Ens  & 41.5          & 78.1          & 66.6          & 29.4          & 53.9          & 53.9          \\
                           & CoTTA-Ens & 28.6          & 76.3          & 64.8          & 60.9          & 73.3          & 60.8          \\
                           & EaTA-Ens  & 28.6          & 71.9          & 58.5          & 61.8          & 72.9          & 58.8          \\
                           & \PLUTO     & \textbf{51.5} & \textbf{84.6} & \textbf{73.8} & \textbf{71.8} & \textbf{76.6} & \textbf{71.7} \\ \hline
\end{tabular}
\label{table:fewshot_Digits}
}

\end{table}
\begin{table}[t]
\caption{\textbf{Results on Office-Home}. Different number of few-shot target samples for training the module selector and tuning the LayerNorm affine parameters.}\vspace{2pt}

\centering
  \resizebox{0.9\columnwidth}{!}{
\begin{tabular}{cccccc|c}
\hline
Data Size                  & Method    & Ar            & Cl            & Pr            & Re            & Avg           \\ \hline
\multirow{4}{*}{128}       & TENT-Ens  & 63.1          & 49.2          & 75.6          & 71.0          & 64.7          \\
                           & CoTTA-Ens & 64.6          & 50.4          & 76.4          & 71.7          & 65.8          \\
                           & EaTA-Ens  & 65.8          & 51.2          & 76.7          & 72.6          & 66.6          \\
                           & \PLUTO     & \textbf{66.0} & \textbf{53.6} & \textbf{79.2} & \textbf{76.3} & \textbf{68.7} \\ \hline
\multirow{4}{*}{64}        & TENT-Ens  & 49.9          & 33.9          & 62.9          & 54.0            & 50.2          \\
                           & CoTTA-Ens & 49.6          & 34.2          & 61.0          & 58.8          & 50.9          \\
                           & EaTA-Ens  & 51.4          & 35.9          & 62.8          & 58.7          & 52.2          \\
                           & \PLUTO     & \textbf{59.7} & \textbf{47.5} & \textbf{71.7} & \textbf{70.2} & \textbf{62.3} \\ \hline
\multirow{4}{*}{16}        & TENT-Ens  & 31.5          & 15.0          & 47.6          & 37.4          & 32.9          \\
                           & CoTTA-Ens & 32.4          & 17.7          & 42.2          & 43.6          & 34.0          \\
                           & EaTA-Ens  & 34.9          & 17.7          & 44.2          & 40.7          & 34.4          \\
                           & \PLUTO     & \textbf{55.9} & \textbf{44.3} & \textbf{67.8} & \textbf{66.4} & \textbf{58.6} \\ \hline
\multirow{4}{*}{Zero-shot} & TENT-Ens  & 27.8          & 11.4          & 44.3          & 34.3          & 29.5          \\
                           & CoTTA-Ens & 29.0          & 13.9          & 38.8          & 39.8          & 30.4          \\
                           & EaTA-Ens  & 30.9          & 13.7          & 41.0          & 37.2          & 30.7          \\
                           & \PLUTO     & \textbf{52.4} & \textbf{40.9} & \textbf{63.9} & \textbf{62.4} & \textbf{54.9} \\ \hline
\end{tabular}
\label{table:fewshot_OfficeHome}
}

\end{table}
\begin{table*}[]
\vspace{-10pt}
\caption{\textbf{Results CIFAR-10C}. We take $M=N=4$ source modules trained on \textit{Snow}, \textit{Frost}, \textit{Fog}, and \textit{Bright}. We employ these models for adaptation on 15 sequential target domains (severity level=1).}
\centering
\resizebox{1\textwidth}{!}{\begin{tabular}{cccccccccccccccc|c}
\hline
           & GN            & SN            & IN            & DB            & FGB           & MB            & ZB            & Snow        & Frost         & Fog           & Bright        & Contrast      & Elastic       & Pixel         & JPEG          & Avg           \\ \hline
CoTTA-Best & 58.0          & 59.3          & 43.9          & 54.6          & 45.7          & 43.0          & 43.8          & 37.0        & 36.1          & 27.5          & 29.0          & 14.3          & 14.4          & 12.9          & 9.7           & 35.3          \\
EaTA-Best  & 55.2          & 57.6          & 45.5          & 66.7          & 58.5          & 66.3          & 69.3          & 69.7        & 72.9          & 70.6          & 69.2          & 67.4          & 62.6          & 68.1          & 52.4          & 63.5          \\
TENT-Best  & 52.4          & 55.6          & 46.0          & 70.1          & 61.4          & 68            & 68.5          & 68.2        & 70.9          & 71.0          & 71.5          & 68.7          & 67.8          & 69.5          & 61.9          & 64.8          \\
CoTTA-Ens  & 58.4          & 61.3          & 42.5          & 57.0          & 45.1          & 44.5          & 45            & 38.5        & 36.0          & 26.7          & 27.4          & 15.4          & 14.9          & 12.8          & 8.2           & 35.6          \\
EaTA-Ens   & 56.2          & 58.5          & 46.7          & 69.2          & 56.6          & 68.3          & 70.4          & 69.2        & 75.5          & 71.5          & 68.9          & 69.9          & 64.7          & 67.1          & 53.8          & 64.4          \\
TENT-Ens   & 53.6          & 57.7          & 45.8          & 71.9          & 59.5          & 68.4          & 68.8          & 69.5        & 72.2          & 70.5          & 69.7          & 69.7          & 67.8          & 69.7          & 62.4          & 65.1          \\
\PLUTO     & \textbf{61.6} & \textbf{63.3} & \textbf{48.7} & \textbf{75.2} & \textbf{64.9} & \textbf{71.8} & \textbf{73.7} & \textbf{73.0} & \textbf{78.9} & \textbf{75.0} & \textbf{74.9} & \textbf{73.2} & \textbf{70.5} & \textbf{72.8} & \textbf{65.0} & \textbf{69.5} \\ \hline
\end{tabular}}
\label{table:forgetting_main}
\end{table*}\vspace{-5pt}\vspace{-5pt}
\begin{table*}[t]
\centering
\caption{\textbf{Results on ImageNet-C.}  \PLUTO adaptation with top $k$ favorable source modules. $k$ stands for the number of used PET source modules in \PLUTO, w.r.t the assigned weight. We adapt 15 target domains (severity level=1) separately.}
\resizebox{1\textwidth}{!}{%
\begin{tabular}{ccccccccccccccccc|c}
\hline
Method                  & k  & GN   & SN   & IN   & DB   & GB   & MB   & ZB   & Snow & Frost & Fog  & Bright & Contrast & Elastic & Pixel & JPEG & Avg  \\ \hline
\multirow{5}{*}{\PLUTO} & 1  & 66.2 & 75.4 & 76   & 74.9 & 74.9 & 77.2 & 75.9 & 75.2 & 76.6  & 76.4 & 76.3   & 76.3     & 75.1    & 77.2  & 76.8 & 75.4 \\
                        & 3  & 69.8 & 78.3 & 78.7 & 77.6 & 76.9 & 79.5 & 77.9 & 77.6 & 79.3  & 79.2 & 78.6   & 79.3     & 77.7    & 79.4  & 79.3 & 77.9 \\
                        & 5  & 72.3 & 78.9 & 79.2 & 78.3 & 77.9 & 80.3 & 78.2 & 78.1 & 80.4  & 80.5 & 79.2   & 80.0     & 78.1    & 79.9  & 79.7 & 78.7 \\
                        & 15 & 72.7 & 79.3 & 79.7 & 78.7 & 78.3 & 80.9 & 78.8 & 78.8 & 81.0  & 80.8 & 79.8   & 80.5     & 78.6    & 80.4  & 80.4 & 79.2 \\
                        & 75 & 72.8 & 79.5 & 80.1 & 78.9 & 78.5 & 81.1 & 79.1 & 79.0 & 81.3  & 81.0 & 80.0   & 80.7     & 78.8    & 80.5  & 80.5 & 79.4 \\ \hline
\end{tabular}
\label{table:selection}
}
\end{table*}

In the main paper, we employ parameter-efficient fine-tuning (PET) in the form of VPT  \cite{jia2022visual,gao2022visual} as our module to demonstrate the effectiveness of our approach. We showcase our method's application to other PET modules in the Appendix Sec.~\ref{sec:supp_PETmethods} to illustrate the generality of our approach. The effectiveness of our approach is evaluated on both stationary and continuous test time adaptation benchmarks. During the testing phase, we evaluate our approach under two scenarios: (1) Test batches are sampled from a stationary distribution (2) Test batches are sampled from a evolving dynamic distribution. \\
\textbf{Datasets.} We demonstrated the efficacy of our approach in handling \underline{stationary distributions} by assessing its performance on the following domain adaptation benchmarks.
\begin{itemize}
    \item \textit{Digits-Five.} \cite{peng2019moment} consists of five-digit datasets: MNIST (MT), MNIST-M (MM), USPS (UP), SVHN (SV), and Synthetic Digits (SY). There are 10 classes corresponding to digits ranging from 0 to 9. We employed four of the domains as sources and reserved the remaining domain for testing.
    \item \textit{Office-Home.} \cite{venkateswara2017deep} The Office-Home dataset consists of four domains: Art (Ar), Clipart (Cl), Product (Pr), and Real World (Re), each containing 65 classes. For evaluating our method in the test-time adaptation scenario, we selected three of these domains as source domains and used the remaining domain as the test data. 
    \item \textit{ImageNet-C.} \cite{hendrycks2019benchmarking} ImageNet-C dataset consists of 15 diverse corruption types applied to images of ImageNet~\cite{deng2009imagenet}. It consists of algorithmically generated corruptions with 5 severity levels. We regard different corruptions with different severity level as different domains. Thus we will have $5\times15=75$ domains at most.

\end{itemize}
In the context of \underline{dynamic test distribution}, we employ the following benchmarks.
\begin{itemize}
    \item \textit{CIFAR-100C.} \cite{hendrycks2019benchmarking} The CIFAR-100~\cite{krizhevsky2009learning} dataset is an image classification dataset consisting of 50,000 training images and 10,000 test images. Based on CIFAR-100, CIFAR-100C \cite{hendrycks2019benchmarking} introduced 15 different types of noise at different severity levels (1 $\rightarrow$ 5). This dataset has since been utilized in various studies on continuous test-time adaptation. 

\end{itemize}

\vspace{-5pt}
\subsection{Baseline Methods}

 Our evaluation includes comparisons with widely recognized single-source test-time adaptation methods, notably TENT \cite{wang2020tent}, CoTTA \cite{wang2022continual}, and EATA \cite{niu2022efficient}. These methods represent the current state-of-the-art (SOTA) in this area. We adopt a setup similar to that in~\cite{ahmed2021unsupervised}, applying each source model individually to specific test domain data. This approach generates \underline{\textbf{Best}} and  \underline{\textbf{Worst}} outcomes, corresponding to the highest and lowest performance achieved among the source models when adapted using a certain method, respectively. 
We also generalize these single-source adaptation methods into multi-source by uniformly averaging pre-softmax logits of all sources' outputs given a target sample to make the prediction. The results of uniformly averaged \underline{\textbf{Ens}}emble of multiple single sources methods are reported.



\subsection{Implementation details}
We use ViT-Base-16~\cite{dosovitskiy2020image} model pretrained on JFT-300M dataset for all our experiments. For all experiments, we use a target batch size of $B=128$, as used by TENT \cite{wang2020tent}. Further details on the experimental settings are provided in Appendix \ref{sec:supp_HP}.

\subsection{Object recognition and digit classification}
\label{sec:exp_singlesource}
We report the results of object recognition on Office-home \cite{venkateswara2017deep} and Digits-Five \cite{peng2019moment} datasets in Table.~\ref{table:Digits}, ~\ref{table:OfficeHome}. We compute the accuracy for each incoming test batch and subsequently present the results by averaging the accuracy values across all the batches. For Office-home dataset, the number of source modules is $N=3$, and we use all $M=N$ (in Alg.~\ref{alg:module_store}) to do the inference on the target domain. A similar setup was used for Digits-Five dataset, except $M=N=4$ here. 
\subsection{Few-shot adaptation} 
Our default setting follows existing studies\cite{wang2020tent} by using 128 few-shot labeled data for each target task, as per the target batch size ($B$), to update the LN affine parameters and the module selector. Our interest lies in examining how \PLUTO's performance changes with varying shots ($U \le B$) compared to other techniques. The outcomes are displayed in Table.~\ref{table:fewshot_Digits} and Table.~\ref{table:fewshot_OfficeHome}.
We observe that \PLUTO consistently outperforms existing TTA methods' uniform ensemble across various amounts of few-shot unlabeled target data. It suggests that \PLUTO can be applied to a wide range of few-shot settings. Even in zero-shot settings \PLUTO can outperform other methods, since \PLUTO benefits from its sample-specific adaptation strategy.

\vspace{-5pt}
\subsection{Forgetting issue}
\label{sec:antiforget}
We extensively experiment with CIFAR-10C dataset to assess our model's performance in the face of dynamic test distributions. Our method's resistance to catastrophic forgetting is showcased by measuring classification accuracy on the original source test set after adapting to each domain \cite{niu2022efficient, song2023ecotta}. For \PLUTO, our ensemble method is employed during domain adaptation, and post-adaptation, each adapted source model is tested on its specific source test set. Regarding the baseline single-source methods and their average ensembles, each model is individually adapted to the new domain and then tested on its respective source test set. The reported accuracy in Table.~\ref{table:forgetting_main} is the average derived from all these individually adapted single-source models. \PLUTO serves an optimal aggregation of source models as well as better preservation of source knowledge.

\vspace{-5pt}
\subsection{Module selection}\vspace{-2pt}
\label{sec:exp_selection}
In previous experiments, we use all the source modules without selection for the target domains as mentioned earlier, regardless of their actual transferability given the target domain. However, if a source domain is too different from the target domain, it is possible that this source task wouldn’t effect the target task positively in general. In order to investigate how \PLUTO performs when less preferable source domains are added, we conduct experiments as follows:\\
On ImageNet-C dataset, we pretrain a module store of $75$ modules($N=75$ in Alg.~\ref{alg:module_store}), each corresponding to a domain. We test \PLUTO by choosing $M=k$ modules based on the module selector weight assignment for each module. The results presented in Table.~\ref{table:selection}, suggest that \PLUTO effectively identifies target samples where less favored source tasks excel and appropriately adjusts sample-specific weights. This ability to rely more on these source models for certain target samples highlights the efficiency of \PLUTO's sample-specific approach.

\vspace{-5pt}
\section{Conclusion}
\PLUTO, our Test Time Adaptation (TTA) algorithm for transformers, adaptively combines several source parameter-efficient tuning (PET) modules during testing.  Our experiments highlight \PLUTO's three key advantages: (1) Its adaptation strategy, which leverages multiple source domains, utilizes diverse domain knowledge for enhanced performance. (2) The sample-specific TTA approach of \PLUTO contributes to its strong few-shot performance. (3) The use of PET modules for minimal parameter updates during test time prevents catastrophic forgetting. On multiple datasets, we demonstrate the power of our proposed approach, especially in a few-shot setting, and with a small number of selected modules.

\vspace{-5pt}
\section*{Impact Statement} This paper presents a framework where a user can effectively adapt their machine learning model to their own needs by downloading modules from a global database. Thus, the outcomes of this work will potentially allow large transformer-based models to be applied to devices with limited computation power, e.g., those at the edge.



\bibliography{main.bib}
\bibliographystyle{icml2024}

\newpage
\appendix
\renewcommand{\thefigure}{\Alph{figure}}
\renewcommand{\thetable}{\Alph{table}}
\renewcommand{\theequation}{\Alph{equation}}
\setcounter{equation}{0}
\setcounter{figure}{0}
\setcounter{table}{0}

\twocolumn[
\section*{Appendix Overview}
\begin{itemize}
    \item Section \ref{sec:supp_SAR} Sharpness-aware domain adaptation
    \item Section \ref{sec:supp_parameterefficiency} Parameter efficiency
    \item Section \ref{sec:supp_PETmethods} Details on different PET methods
    \item Section \ref{sec:supp_HP} Hyperparameters
    \item Section \ref{sec:supp_corruption} Details of dataset corruptions
    \item Section \ref{sec:theory} Theoretical insights
\end{itemize}
]

\section{Sharpness-aware domain adaptation}
\label{sec:supp_SAR}
Following \cite{niu2023towards}, in addition to minimizing the weighted pseudo-label entropy with Eqn.~\ref{eq:SAM_GradientApprox}, we also seek to filter the samples that of large entropy value. As filtering samples based on their gradient norms directly is impractical, our approach focuses on examining the connection between entropy loss and gradient norms. We aim to eliminate samples with large gradients by considering their entropy values. The entropy is determined by the number of output classes, denoted as $K$, and it falls within the range $(0, \ln K)$ for various models and datasets. Consequently, selecting a threshold for filtering samples based on their entropy is a more manageable task. The threshold $E_0$ is set to $0.4 \ln K$ by following EATA \cite{niu2022efficient}. $\rho$ in Eqn.~\ref{eq:SAMsol} is set by the default value $5\times10^{-2}$ in \cite{foret2020sharpness}. We provide a overall view of the sharpness-aware DA applied in our method, PLUTO.

\begin{algorithm}[]
\caption{Sharpness-Aware Test-Time Entropy Minimization}
\label{alg:SA_entropyminimization}
\begin{algorithmic}[1]
\State \textbf{Input:} $t$-th target samples $D_T^{(t)}=\{x_i^{(t)}\}_{i=1}^B$, $j$-th source model $f^j_\lambda$ with tunable LN affine parameters $\lambda$, step size $\eta>0$, neighbourhood size $\rho > 0$, entropy threshold $E_0$,
\State \textbf{Output: updated LN parameter $\lambda^*$}
\For {Each $x_i^{(t)}$ in $D_T^{(t)}$}
    \State Predict pseudo-label $\hat{y}_i^{(t)}=f^j_\lambda(x_i^{(t)})$
    \State Compute entropy $\mathcal{E}_i^{(t)}$ 
    \Statex \hspace{\algorithmicindent} $ \mathcal{E}_i^{(t)}(x_i^{(t)};\lambda)=-\sum_{c=1}^K \hat{y}_{ic}^{(t)}\log(\hat{y}_{ic}^{(t)})$

    \If {$\mathcal{E}_i^{(t)}>E_0$}
        \State \textbf{continue} \textcolor{gray}{//filter the sample with large entropy}
    \EndIf
    \State Compute gradient $\mathbf{v}_i=\nabla_\lambda \mathcal{E}_i^{(t)}(x_i^{(t)};\lambda)$
    \State Compute $\mathbf{\epsilon}^*_i(\lambda)=\rho \text{sgn} (\mathbf{v}_i)\frac{(\mathbf{v}_i)}{\|(\mathbf{v}_i)\|_2}$
    \State Compute gradient approximation: 
    \Statex \hspace{\algorithmicindent}$\mathbf{g}_i=\nabla_\lambda \mathcal{E}_i^{(t)}(x_i^{(t)};\lambda)|_{\lambda+\mathbf{\epsilon}^*_i(\lambda)}$
    \State Update $\lambda \leftarrow \lambda - \eta \mathbf{g}_i$
\EndFor
\end{algorithmic}
\end{algorithm}

By applying Alg.~\ref{alg:SA_entropyminimization}, we minimize Eqn.~\ref{eq:LNentropy} with Eqn.~\ref{eq:SAM_GradientApprox}. On Office-Home dataset, we provide an ablation study of how much each component contribute to the performance in Table.~\ref{table:supp_OfficeHome} (Here, we apply VPT as the PET module, same as in the main paper Table.~\ref{table:OfficeHome}). We demonstrate the three components of PLUTO, namely, module selection, sharpness-aware pseudo-label entropy minimization (SA), and high-entropy sample filtering, through ablation experiments, highlighting their contributions to domain adaptation. It is evident that SA and Filter, in addition to module selection, contribute to the improved performance of PLUTO.
\begin{table}[]

  \centering
        \caption{\textbf{Results on Office-Home}. We train the source modules using 3 datasets to perform inference on the remaining dataset. }
  \resizebox{\columnwidth}{!}{%
    \begin{tabular}{c ccccc c}
      \toprule
    \textbf{Source} & \textbf{Method} & \textbf{Ar} & \textbf{Cl} & \textbf{Pr} & \textbf{Re} & \textbf{Avg}\\  
      \cmidrule(lr){1-6} \cmidrule(lr){7-7} 
      \multirow{2}{*}{\makecell{Baseline\\(In Main paper)}} & TENT-Ens  & 63.1 & 49.2 & 75.6 & 71.0 & 64.7\\
        &  PLUTO   & \textbf{66.0} & \textbf{53.6} & \textbf{79.2} & \textbf{76.3} & \textbf{68.7}\\
      \cmidrule(lr){1-6} \cmidrule(lr){7-7} 
      \multirow{3}{*}{\makecell{Variants of \\PLUTO}} & Module selection & 63.4 & 50.1 & 76.1 & 73.4 &  65.8\\
        & + SA & 65.1 & 52.9 & 78.3 & 75.6 & 67.9\\
        & + Filter & 63.7 & 51.5 & 76.8 & 74.1 & 66.3\\
      \bottomrule
    \end{tabular}
  }

  \centering
  \label{table:supp_OfficeHome}
\end{table}

\section{Parameter efficiency}
\label{sec:supp_parameterefficiency}
\begin{table}[]
\centering\caption{Number of parameters for different models' scales and their corresponding PET module size in our setting. ViT-T/S/B/L stands for "Tiny, Small, Base, Large", corresponding to different pretrained ViT sizes. The bolded numbers are the size of the PET modules we applied in our experiments. We highlighted the two size of PET modules in our experiments.}
\small{
  \resizebox{\columnwidth}{!}{\begin{tabular}{lllll}
        & ViT-T & ViT-S & ViT-B & ViT-L \\ \hline
Full model    & 5,543,716 & 21,704,164    &   85,875,556    &   303,404,132 \\
Adapter  &   58,564    &     116,932  &  \textbf{233,668}     &    417,984  \\     
VPT     &    37,732   &   75,364    &    \textbf{150,628}   &   299,108   \\
Header &  19,300     &  38,500     &  76,900     &   102,500  
\end{tabular}}}\label{table:supp_num_param}
\end{table}

Parameter-efficient tuning (PET) methods align naturally with model ensemble techniques\cite{lester2021power}, particularly in terms of parameter efficiency. In contrast to other models where an ensemble of $N$ models results in $N$ times more model parameters, an ensemble of $N$ distinct PET modules merely entails $N$ times more modules. This is because all models to be ensembled share the same pre-trained model that conditions the PET modules. Therefore, although PLUTO is model ensemble test-time adaptation approach, the additional model parameters introduced by the ensemble are only the PET modules, i.e. $N \times |\theta|$. In our approach, we train an attention-based module selector $\mathcal{G}$ which includes four projection layers and two layer norms. It required $d\times d'_x +d'_x\times d'+v\times d'_l + d'_l \times d'+ 4d' $ parameters ($\approx$ 0.8M in our experiment). It only required about less than 1\% of a pretrained ViT-base model ($\approx$ 86M, Table.~\ref{table:supp_num_param}).  
\section{Details on different PET methods}

\label{sec:supp_PETmethods}
\subsection{Visual-Prompt Tuning (VPT)}
With a pre-trained Transformer (ViT) model as our starting point, we introduce a set of $p$ continuous embeddings in the input space, each of dimension $d$, referred to as "prompts." During fine-tuning, only the prompts specific to the task are updated, while the Transformer backbone remains frozen. We applied VPT-Shallow\cite{jia2022visual} as following:

\textbf{VPT-Shallow.} Prompts are inserted into the first Transformer layer only.
Each prompt token is a learnable $d-$ dimensional vector. A module, which is a collection of $p$ (which is the prompt length) prompts is denoted as $\mathbf{P}=\{\mathbf{v} \in \mathbb{R}^d|k\in \mathbf{N}, 1\le k \le p\}$, the shallow-prompted ViT is:
\begin{align}
    & [x_1, \mathbf{Z}_1, \mathbf{E}_1]=\textcolor{blue}{L_1}([\textcolor{blue}{x_0}, \mathbf{\textcolor{red}{P}}, \mathbf{E}_0]) \\
    & [x_i, \mathbf{Z}_i, \mathbf{E}_i]=\textcolor{blue}{L_i}([x_{i-1}, \mathbf{Z}_{i-1}, \mathbf{E}_{i-1}]), i=2,\cdots, l\\
    & \mathbf{y} = \textcolor{red}{\text{Head}}(x_l),
\end{align}
where $\mathbf{Z}_i \in \mathbb{R}^{p\times d}$ represents the features computed by the $i$-th Transformer layer, and $[x_i, \mathbf{Z}_i, \mathbf{E}_i]\in \mathbb{R}^{(1+p+P)\times d}$ ($P$ is the number of patches that a 2D image input is divided into). The color above indicate \textcolor{red}{learnable} and \textcolor{blue}{frozen} parameters, respectively. For ViTs, $x_i$ is invariant to the location of prompts since they are inserted after positional encoding. The overall parameter count for Adapters in an $l$-layer Transformer can be calculated as $|\theta|= p\times d.$

\subsection{Adapter} 
In the conventional configuration, a transformer model incorporates two Adapters per layer \cite{baevski2018adaptive}. Each Adapter layer is composed of $2 \times k \times d$ parameters, accounting for both the down and up-projection matrices. Here, $k$ represents the input dimension size, while $d$ refers to the bottleneck dimension of the Adapter. The overall parameter count for Adapters in an $l$-layer Transformer can be calculated as $|\theta|=2 \times l \times 2 \times k \times d.$

\subsection{Results for different PET modules}
Following the setup in main paper Sec.~\ref{sec:exp_singlesource}, we replace VPT with Adapters as the PET module, and demonstrate the stationary distribution adaptation results in Table.~\ref{table:supp_Adapter_OfficeHome} and Table.~\ref{table:supp_Adapter_Digits}. In can be concluded that our method, PLUTO, remains good performance across different PET modules. 

\begin{table}[t]
\vspace{-10pt}
  \centering
      \caption{\textbf{Results on Office-Home}. We train the source modules using 3 datasets to perform inference on the remaining dataset ($M=N=3$). The difference here is that we applied Adapter instead of VPT in the main paper.}
  \resizebox{0.9\columnwidth}{!}{%
\centering
\begin{tabular}{ccllll|l}
\hline
\textbf{Source}         & \textbf{Method} & \multicolumn{1}{c}{\textbf{Ar}} & \multicolumn{1}{c}{\textbf{Cl}} & \multicolumn{1}{c}{\textbf{Pr}} & \multicolumn{1}{c|}{\textbf{Re}} & \multicolumn{1}{c}{\textbf{Avg}} \\ \hline
\multirow{5}{*}{Single} & TENT-Best       & 65.6                            & 52                              & 77.9                            & 72.1                             & 66.9                             \\
                        & TENT-Worst      & 47.5                            & 47.1                            & 57.6                            & 65.3                             & 54.4                             \\
                        & CoTTA-Best      & 65.3                            & 50.9                            & 75.8                            & 70.9                             & 65.7                             \\
                        & CoTTA-Worst     & 46.6                            & 46.2                            & 56.1                            & 63.8                             & 53.2                             \\
                        & EaTA-Best       & 65.5                            & 51.8                            & 76.6                            & 72.0                             & 66.5                             \\
\multicolumn{1}{l}{}    & EaTA-Worst      & 46.7                            & 46.3                            & 55.2                            & 61.9                             & 52.5                             \\ \hline
\multirow{4}{*}{Multi}  & TENT-Ens        & 63.1                            & 49.2                            & 76.6                            & 72.9                             & 65.5                             \\
                        & CoTTA-Ens       & 65.4                            & 49.9                            & 77.2                            & 71.8                             & 66.1                             \\
                        & EaTA-Ens        & 65.2                            & 50.7                            & 76.3                            & 71.6                             & 66.0                             \\
                        & \PLUTO (Ours)   & \textbf{65.7}                   & \textbf{55.4}                   & \textbf{79.1}                   & \textbf{77.1}                    & \textbf{69.3}                    \\ \hline
\end{tabular}
  }
  \centering
  \label{table:supp_Adapter_OfficeHome}
\end{table}

\begin{table}[t]
\vspace{-10pt}
  \centering
      \caption{\textbf{Results on Digits}. We train the source modules using 4 digits datasets to perform inference on the remaining dataset ($M=N=4$). The difference here is that we applied Adapter instead of VPT in the main paper.}
  \resizebox{\columnwidth}{!}{%
\begin{tabular}{ccccccc|c}
\hline
\textbf{Source}         & \textbf{Method} & \textbf{MM}   & \textbf{MT}   & \textbf{UP}   & SV            & \textbf{SY}   & \textbf{Avg}  \\ \hline
\multirow{5}{*}{Single} & TENT-Best       & 54.5          & 97.6          & 83.7          & 86.6          & 94.5          & 83.4          \\
                        & TENT-Worst      & 16.4          & 54.0          & 59.2          & 11.7          & 15.7          & 31.4          \\
                        & CoTTA-Best      & 53.5          & 98.5          & 85.8          & 84.8          & 94.5          & 83.4          \\
                        & CoTTA-Worst     & 16.2          & 53.9          & 58.6          & 10.3          & 16.3          & 31.1          \\
                        & EaTA-Best       & 55.4          & 97.4          & 83.6          & 85.9          & 92.8          & 83.0          \\
                        & EaTA-Worst      & 16.9          & 51.7          & 57.9          & 8.7           & 16.5          & 30.3          \\ \hline
\multirow{4}{*}{Multi}  & TENT-Ens        & 61.8          & 98.8          & 85.8          & 53.2          & 77.1          & 75.3          \\
                        & CoTTA-Ens       & 56.2          & 98.6          & 85.5          & 87.0          & 95.3          & 84.5          \\
                        & EaTA-Ens        & 56.3          & 96.7          & 83.8          & 86.4          & 94.5          & 83.5          \\
                        & \PLUTO (Ours)   & \textbf{64.0} & \textbf{99.1} & \textbf{89.9} & \textbf{87.3} & \textbf{95.5} & \textbf{87.2} \\ \hline
\end{tabular}
  }
  \centering
  \label{table:supp_Adapter_Digits}
\end{table}

\section{Hyperparameters}
\label{sec:supp_HP}

The experimental setup of the attention-based module selector $\mathcal{G}$ is listed as in Table.~\ref{table:supp_HP_G}.
Table.~\ref{table:supp_HyperParam_PET} summarizes the optimization configurations we used. Implementation details for each tuning method on source/target tasks are also included.

\begin{table*}[htbp]
\centering
\caption{Hyperparameters of the module selector $\mathcal{G}$ for different datasets.}
\begin{tabular}{l|l|llll}
\hline
Model                     & Hyper-param & CIFAR100-C & ImageNet-C & Office-Home & 5-Digits \\ \hline
\multirow{4}{*}{ViT-Base} & $d'_x $       & 64        & 128        & 32          & 32       \\
                          & $d'_l $       & 64        & 128        & 32          & 32       \\
                          & $d'$          & 64        & 128        & 64          & 32       \\
                          & Dropout \%   & 0         & 0          & 0           & 50       \\ \hline

\end{tabular}

\label{table:supp_HP_G}
\end{table*}
\begin{table*}[]
\centering
\caption{Hyperparameters and implementation details for each tuning methods.}
\begin{tabular}{l|cc}
\hline
                       & Full, Adapter                  & VPT                                               \\ \hline
Optimizer              & AdamW                          & SGD                                               \\
Optimizer momentum     & \textcolor{gray}{N/A}          & 0.9                                               \\
$base\_lr$ search range & \{0.001, 0.0001, 0.0005, 0.005\} & \{50., 25., 10., 5., 2.5, 1.,0.5, 0.25, 0.1, 0.05\} \\
Weight decay range     & \multicolumn{2}{c}{\{0.01, 0.001, 0.0001, 0.0\}}                                     \\
LR schedule            & \multicolumn{2}{c}{cosine decay}                                                   \\
Warm up epochs         & \multicolumn{2}{c}{10}                                                             \\
Total epochs           & \multicolumn{2}{c}{100}                                                            \\ \hline
\end{tabular}

\label{table:supp_HyperParam_PET}
\end{table*}
\section{Details of dataset corruptions}
\label{sec:supp_corruption}
\begin{figure*}[htbp]
  \centering
  \includegraphics[width=0.87\textwidth]{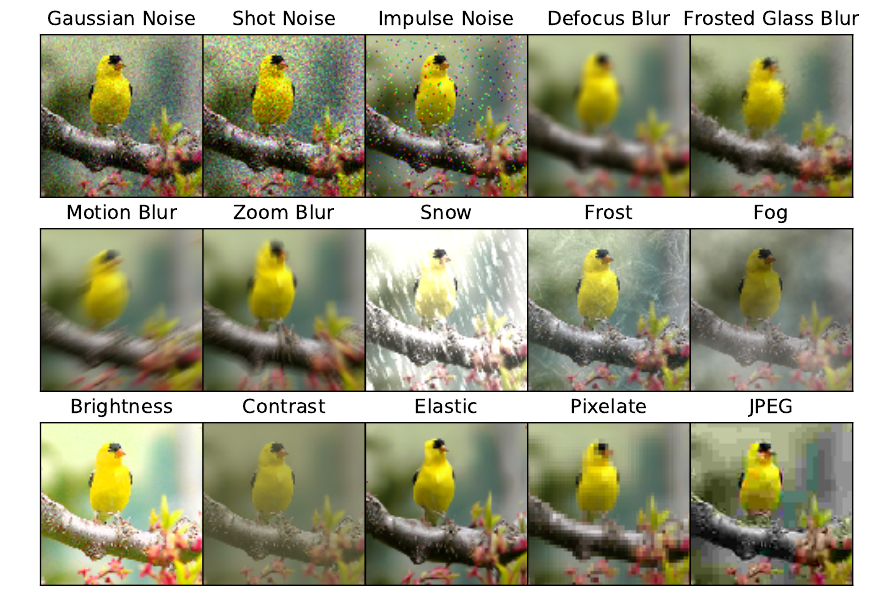}
  \caption{Examples of each corruption type in the image corruptions benchmark. While synthetic, this set of corruptions aims to represent natural factors of variation like noise, blur, weather, and digital imaging effects.} \label{fig:supp_data_noise}
\end{figure*}

We summarize these corruptions types by example in Fig.~\ref{fig:supp_data_noise}. The order of these corruptions is same as the order in Table.~\ref{table:forgetting_main} and Table.~\ref{table:selection}
\clearpage
\section{Theoretical Insights}
\label{sec:theory}

\PLUTO's goal is to determine the ideal weights $\{w(x)^j\}_{j=1}^N$ for every source, combining them to create the target predictor. We will demonstrate that, given logical assumptions about the source and target distributions, there is a straightforward selection of the target predictor. This selection can match or surpass the performance of the best source model when applied directly to the target data.\\
Formally, consider $L$ as a convex loss function that maps a pair consisting of a model-predicted label and the actual ground-truth label to a scalar value. Let the expected loss over the $k$-th source distribution $Q_S^k$, utilizing the source predictor $\theta$, be denoted as $\mathcal{L}=\mathbb{E}_x [L(\theta(x),y)]$, which can also be expressed as $\int_x L(\theta(x),y)Q_s^k(x)dx$. If the target distribution can be expressed by an linear combination of source distributions, i.e. $Q_T(x)=\sum_{k=1}^N \lambda_k Q_S^k(x), \lambda_k\ge 0, \sum_{k=1}^N\lambda_k=1,$ and the source predictors are trained/finetuned to be optimal: $$\theta_S^k=\text{argmin}_\theta \mathcal{L}(Q_S^k,\theta),\forall 1\le k\le N,$$ then the target predictor can be experssed as $$\theta_T(x)=\sum_{k=1}^N \frac{\lambda_k Q_S^k(x)}{\sum_{j=1}^N\lambda_j Q_S^j(x)}\theta^k_S(x).$$
By applying Lemma 1,2 in \cite{ahmed2021unsupervised}, we can know that the pseudo-label loss (Eqn.~\ref{eq:PLentropy}) and the supervised loss are both less than or equal to the loss induced by the best source predictor, i.e. $$\mathcal{L}(Q_T,\theta_T)\le \min_{1\le j\le N} \mathcal{L}(Q_T, \theta_S^j).$$
Which is consistent with the results shown in Table.~\ref{table:Digits},~\ref{table:OfficeHome}.

\end{document}